# A Line-Point Unified Solution to Relative Camera Pose Estimation*

Ashraf Qadir, Jeremiah Neubert

*Abstract*— In this work we present a unified method of relative camera pose estimation from points and lines correspondences. Given a set of 2D points and lines correspondences in three views, of which two are known, a method has been developed for estimating the camera pose of the third view. Novelty of this algorithm is to combine both points and lines correspondences in the camera pose estimation which enables us to compute relative camera pose with a small number of feature correspondences. Our central idea is to exploit the tri-linear relationship between three views and generate a set of linear equations from the points and lines correspondences in the three views. The desired solution to the system of equations are expressed as a linear combination of the singular vectors and the coefficients are computed by solving a small set of quadratic equations generated by imposing orthonormality constraints for general camera motion. The advantages of the proposed method are demonstrated by experimenting on publicly available data set. Results show the robustness and efficiency of the method in relative camera pose estimation for both small and large camera motion with a small set of points and line features.

I. INTRODUCTION

Accurate and efficient camera pose estimation is essential to a number of applications such as structure from motion, robot navigation and simultaneous localization and mapping. One key step is to estimate the relative position and orientation of the views using the geometric constraints arising from a 3D object and its corresponding 2D images in multiple views. Most existing methods use images of scene features such as corner points [1, 2, 3] or lines [4, 5] with points being the primary feature of interest. However, almost all of these methods use either points or lines but do not handle both. Using either lines or points does not allow one to take full advantages of the information available in the images. While natural scenes are rich in feature points, man-made environment such as office interiors or city landscapes can consists of texture less planar surfaces where there are few point features that can be reliably detected. On the other hand, such environments consists of large number of visible lines. A system that can handle both lines and points, either independently or combined has an advantage to be useful in a wide range of applications. While visual SLAM systems mostly deal with sequential images with small camera motions, structure from motion methods often require to deal with an unordered set of images with large camera motions.

In this work we propose an algorithm that gives a unified linear approach that can deal with a mixture of points and lines for relative pose estimation using three views. Given a set of lines and points correspondences between views, the algorithm exploits the tri-linearity relationship between the views to generate a set of constraint equations from the points and lines correspondences. The desired solution to the system of equations are expressed as a linear combination of the singular vectors and the coefficients are computed by solving a small set of quadratic equations generated by imposing orthonormality constraints for general camera motion.

The proposed method considers the image sequence is generated from a monocular camera and equation representations used are adopted to three views. However, the method can be easily extended to stereo systems with immediate advantage that only two view are required. It also generates an added set of constraint equations when using points and line correspondences in all 4 views.

In theory given two known camera pose, it is possible to reconstruct 3D lines and points from the corresponding images and then estimate the third camera pose from the 3D to 2D points and lines correspondences. However, the quality of the camera pose estimation is critically dependent on the accuracy of the reconstructed 3D points and lines. It is especially critical with narrow base line and forward motion [4].

The major contribution of this paper is the development of a unified framework for relative camera pose estimation combining points and lines. The algorithm is robust, efficient and can handle a wide range of line and point combination. The intended use of the algorithm is with robust hypothesize-and-test frameworks but is also suitable for scenarios with a large number of points and lines. Using both points and line features mean the proposed method is equally applicable in natural as well as manmade urban and indoor environments.

II. LITERATURE REVIEW

While line based structure from motion has received considerable attention, most of the existing systems that are robust and computationally efficient are points based systems. A primary reason is that points provide stronger constraints than lines as shown in [6]. For instance, the 5 points minimal solution exists [1] for two view reconstruction for a calibrated camera. However, lines do

*Research supported partially by ND EPSCoR through NSF grants #EPS-0814442, #IIA-1355466 and the University of North Dakota

Ashraf Qadir is with RFA Engineering, Minnesota, USA (phone: 1-701-213-8094; e-mail: aqadir@ rfamec.com).

Jeremiah Neubert is with Mechanical Engineering Department, University of North Dakota, Grand Forks, North Dakota.(email: jeremiah.neubert@und.edu)

not put such constraints on the camera pose and at least 3 images are required for camera placement derivation. As a result, a common approach is to use multifocal tensors [6, 7] for structure from motions from lines. The disadvantage is that a large number of lines in three views are required making it less desirable in a hypothesis-and-test framework for motion estimation. Yet there are situations where using lines are beneficial or required for a successful structure from motion system.

Important breakthrough in solving minimal problems for points [1, 2] led to the development of efficient point based camera pose estimation and reconstruction methods. A few examples are visual odometry methods [8, 9, 10] or urban reconstruction [11, 12] that uses the 5 points algorithm inside a RANSAC framework for relative camera pose estimation. These methods rely on the assumption that many points can be reliably detected and localized. However, stability of 5 points algorithm decreases for forward motions as well as large depth of the features. For robotics application, often additional sensors (IMU, odometer etc.) and constraints are used along with vision for relative pose estimation [13].

One of the earlier methods for line based 3D structure and camera motion estimation is described in [14]. In order to obtain an estimation of the scene structure and the motion of the camera together, an objective function is defined that minimizes the re-projection error in the image plane. The method is robust to variable end points across views. However, they use iterative nonlinear minimization which requires initialization. The system also requires at least 6 line correspondences over three views. A similar method is presented in [15] for urban environment.

A detail investigation on 3D line representation, triangulation and bundle adjustment is performed in [16] where the camera pose estimation was performed using trifocal tensor. An efficient algorithm for relative camera pose estimation using two stereo image pairs (four images) is presented in [4]. The algorithm uses 2 or 3 line correspondences in four images of two stereo image pairs to estimate the relative camera pose between two stereo frames in a RANSAC framework. That paper contains elements of ideas contained in the present paper which extends the idea to monocular camera with a combine points and line based relative pose estimation as well as large camera motions.

Focusing on indoor and/or urban application, one category of line based SFM from lines algorithms proposed in [17, 18] based on the assumptions such as existence of three orthogonal directions or primitive configurations such as parallel and orthogonal lines. In [18], three dominant directions are computed for camera pose recovery. All lines in an image are used to find the dominant direction and the method fails if a dominant direction cannot be detected. On the other hand, the method presented in [5] requires a primitive configuration of two parallel lines and an orthogonal line to estimate the relative camera pose in a RANSAC framework. As a result the method is not suitable for environment with few or no primitive configurations among lines.

In literature, there are a few methods present that combines line and points together for camera pose recovery or structure from motion. Once such method is presented in [6] that involves generating a set of linear equations from the point and line correspondences in three views and solve for the trifocal tensor. Camera poses for the second and third views are then computed from the estimated trifocal tensor. A linear solution followed by an iterative refinement was presented. The method presented in this work is foreshadowed by the aforementioned work and extends to a robust and efficient relative camera pose estimation.

### III. OUR APPROACH TO RELATIVE POSE ESTIMATION

Let the three cameras are denoted as $\mathbf{P_1}, \mathbf{P_2}$, and $\mathbf{P_3}$. Assuming that the intrinsic parameter matrix $\mathbf{K}$ is known, the first two cameras can be parameterized as $\mathbf{P_1} = [\mathbf{I} | \mathbf{0}]$ and $\mathbf{P_2} = [\mathbf{R_0} | \mathbf{t_0}]$ where $\mathbf{R_0}$ is the known rotation matrix and $\mathbf{t_0}$ is the known translation vector of the second camera relative to the first camera. Then the camera matrix of the third camera relative to the first camera is given by $\mathbf{P_3} = [\mathbf{R} | \mathbf{t}]$.

#### A. Problem formulation from 2D lines correspondences in Three Views

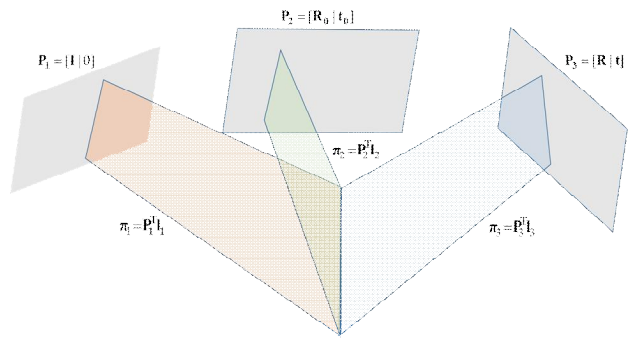

Fig. 1: Geometry of 3D line formation. The back projected planes joining three camera centers and the three corresponding lines intersect in a 3D line.

Let $\mathbf{L}$ be a 3D line and its projection on three image planes are $\mathbf{l_1}, \mathbf{l_2}$ and $\mathbf{l_3}$. The back projected planes through the three camera centers containing the 3D line L and imaged as $\mathbf{l_1}, \mathbf{l_2}$ and $\mathbf{l_3}$ are given by $\boldsymbol{\pi_1} = \mathbf{P_1^T l_1}$, $\boldsymbol{\pi_2} = \mathbf{P_2^T l_2}$ and $\boldsymbol{\pi_3} = \mathbf{P_3^T l_3}$. Together the three equation can be stacked in a 4x3 matrix form as

$$\mathbf{M} = \begin{bmatrix} \mathbf{l_1} & \mathbf{R_0^T l_2} & \mathbf{R^T l_3} \\ \mathbf{0} & \mathbf{t_0^T l_2} & \mathbf{t^T l_3} \end{bmatrix} \quad (1)$$

Since the three back projected planes intersect at the 3D line $\mathbf{L}$, there are two independent 3D points $\mathbf{X_1}$ and $\mathbf{X_2}$ on the line $\mathbf{L}$ that satisfy the point-plane incident relations $\mathbf{X_i^T} \boldsymbol{\pi_j} = 0$, for $i = 1,2$ and $j = 1,2,3$. As a result the matrix M is a rank 2 matrix with 1D null space and we can consider that the first column $m_1$ is a linear combination of the last two columns $m_1 = am_2 + bm_3$.

Similar to [], we perform a two-step QR decomposition of the matrix M with householder rotation to obtain a matrix of the form

$$\mathbf{M}' = \begin{bmatrix} \times & \times & \times \\ 0 & \times & \times \\ 0 & 0 & f_1(\mathbf{R},\mathbf{t}) \\ 0 & 0 & f_2(\mathbf{R},\mathbf{t}) \end{bmatrix} \quad (2)$$

Where $f_1(\mathbf{R},\mathbf{t})$ and $f_2(\mathbf{R},\mathbf{t})$ are two affine functions of the parameters $\mathbf{R},\mathbf{t}$. The matrix $\mathbf{M}'$ is also of rank 2 since the rank of a matrix is preserved by elementary matrix operation. From the above matrix it readily follows that

$$f_i(\mathbf{R},\mathbf{t}) = 0; i = 1,2 \quad (3)$$

### B. Problem formulation from Points Correspondences in Three Views

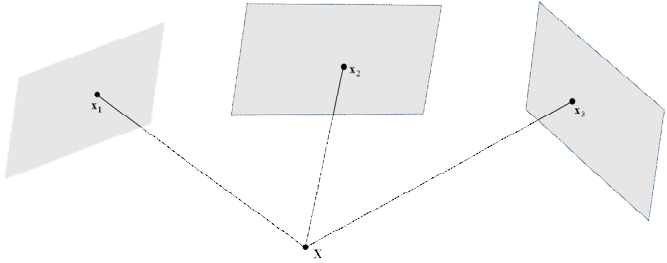

Fig 2. The correspondence $\mathbf{x_1} - \mathbf{x_2} - \mathbf{x_3}$ arises from the image of a 3D point $\mathbf{X}$ in space

Let $\mathbf{X}$ be a 3D point in space and its projection on three image planes are $\mathbf{x_1}, \mathbf{x_2}$, and $\mathbf{x_3}$ respectively. The tri-linearity constraint between the three views gives rise to four independent equations involving the three corresponding points and the three camera matrices [19, 20]. Using tensor notation, the four equations can be written as

$$x_1^k(x_2^1 x_3^1 T_k^{33} - x_3^1 T_k^{13} - x_2^1 T_k^{31} + T_k^{11}) = 0 \quad (4)$$

$$x_1^k(x_2^1 x_3^2 T_k^{33} - x_3^2 T_k^{13} - x_2^1 T_k^{32} + T_k^{12}) = 0 \quad (5)$$

$$x_1^k(x_2^2 x_3^1 T_k^{33} - x_3^1 T_k^{23} - x_2^2 T_k^{31} + T_k^{21}) = 0 \quad (6)$$

$$x_1^k(x_2^2 x_3^2 T_k^{33} - x_3^2 T_k^{23} - x_2^2 T_k^{32} + T_k^{22}) = 0 \quad (7)$$

where $T_i^{jk} = a_i^j b_4^k - a_4^j b_i^k$ is the tri-focal tensor, and $\mathbf{a_i}, \mathbf{b_i}$ are the columns of the camera matrices $\mathbf{P_2} = [\mathbf{a_1} \ \mathbf{a_2} \ \mathbf{a_3} \ \mathbf{a_4}]$ and $\mathbf{P_3} = [\mathbf{b_1} \ \mathbf{b_2} \ \mathbf{b_3} \ \mathbf{b_4}]$. For better numerical stability, the points are normalized following the methods mentioned in [19]. If $\mathbf{F}, \mathbf{G},$ and $\mathbf{H}$ represents three coordinate transformations in three images and the points are transformed as $\hat{x}_1^j = F_i^j x_1^i$, $\hat{x}_2^j = G_i^j x_2^i$ and $\hat{x}_3^j = H_i^j x_3^i$, then the transformed trifocal tensor is written as

$$\hat{T}_i^{jk} = (F^{-1})_s^r G_s^j H_t^k T_r^{st} \quad (8)$$

Then the new set of equations become

$$\hat{x}_1^k(\hat{x}_2^1 \hat{x}_3^1 \hat{T}_k^{33} - \hat{x}_3^1 \hat{T}_k^{13} - \hat{x}_2^1 \hat{T}_k^{31} + \hat{T}_k^{11}) = 0 \quad (9)$$

$$\hat{x}_1^k(\hat{x}_2^1 \hat{x}_3^2 \hat{T}_k^{33} - \hat{x}_3^2 \hat{T}_k^{13} - \hat{x}_2^1 \hat{T}_k^{32} + \hat{T}_k^{12}) = 0 \quad (10)$$

$$\hat{x}_1^k(\hat{x}_2^2 \hat{x}_3^1 \hat{T}_k^{33} - \hat{x}_3^1 \hat{T}_k^{23} - \hat{x}_2^2 \hat{T}_k^{31} + \hat{T}_k^{21}) = 0 \quad (11)$$

$$\hat{x}_1^k(\hat{x}_2^2 \hat{x}_3^2 \hat{T}_k^{33} - \hat{x}_3^2 \hat{T}_k^{23} - \hat{x}_2^2 \hat{T}_k^{32} + \hat{T}_k^{22}) = 0 \quad (12)$$

Similar to the equations for line correspondences, we re-arrange the above equations to generate four independent equations of the form

$$f_j(\mathbf{R},\mathbf{t}) = 0; j = 1,2,3,4 \quad (13)$$

### C. Liner Solution

Every corresponding triplet of points $\mathbf{x_1}, \mathbf{x_2}, \mathbf{x_3}$ contributes four linearly independent equations. Similarly every corresponding triplet of lines $\mathbf{l_1}, \mathbf{l_2}, \mathbf{l_3}$ contributes two linearly independent equations. The equations can be stacked to construct a system of the form

$$\mathbf{Av} = \mathbf{b} \quad (14)$$

where $\mathbf{v} = [\mathbf{r_1^T r_2^T r_3^T t^T}]$ and $\mathbf{r_1}, \mathbf{r_2}, \mathbf{r_3}$ are the columns of the rotation matrix R while $t$ is the translation vector. $\mathbf{A}$ is a $n \times 12$ matrix, generated by arranging the coefficients from the equations in (3) and (13). If the number of point triplets is denoted as $n_1$ and the number of line triplets is denoted as $n_2$, then the number of rows is $n = 4*n_1 + 2*n_2$. Since $\mathbf{v}$ is a 12 element vector, we need at least 12 constraint equations generated from the lines or points correspondences. However, such a system suffers from in the presence of noise and the solution can be arbitrarily far from a rigid motion since orthonormality constraints are ignored.

### D. Solution as a Linear Combination of Singular Vectors with Orthonormality Constraints

Once the matrix A is constructed, the solution of the linear system lies in the null space or the kernel of A. In noise free cases the rank of the matrix A is one (because of scale ambiguity) when the number of independent linear equations $n \geq 12$ generated from at least three points correspondences or six line correspondences or a combination of points and lines (assuming points and lines are all independent). However, the matrix A can be rank deficient because of noise or nearly parallel motion (principal axes are parallel). In addition the solution is also required to satisfy the geometric constraints among the elements of $\mathbf{v}$. Similar to

[..] we express the desired solution as a linear combination of the last k singular vectors of the matrix A, denoted as $\mathbf{v_i}$

$$\begin{bmatrix} \mathbf{r_1} \\ \mathbf{r_2} \\ \mathbf{r_3} \\ \mathbf{t} \end{bmatrix} = \beta_1 \begin{bmatrix} | \\ | \\ \mathbf{v_1} \\ | \\ | \end{bmatrix} + \ldots + \beta_k \begin{bmatrix} | \\ | \\ \mathbf{v_k} \\ | \\ | \end{bmatrix} \quad (15)$$

The problem is then reduced to determining the coefficients $\beta_1, \beta_2, \ldots, \beta_k$ of the above linear combination, subject to orthonormality constraints:

$$\|\mathbf{r_1}\|^2 = 1, \quad \|\mathbf{r_2}\|^2 = 1, \quad \|\mathbf{r_3}\|^2 = 1 \quad (16)$$

$$\mathbf{r_1^T r_2} = 0, \quad \mathbf{r_2^T r_3} = 0, \quad \mathbf{r_3^T r_1} = 0 \quad (17)$$

Substituting ($\mathbf{R}, \mathbf{t}$) from equation (15) in equation (16) and (17) we get 6 polynomials of degree two in the k variables $\beta_1, \beta_2, \ldots, \beta_k$. This system of polynomial equations will have no solution in the general noisy case and we need to resort to a principled "least squares" approach to extract the solution. The degree of the system can be reduced by replacing the unit norm constraint by the equal norm constraints:

$$\|\mathbf{r_1}\|^2 - \|\mathbf{r_2}\|^2 = 0, \text{ and } \|\mathbf{r_2}\|^2 - \|\mathbf{r_3}\|^2 = 0 \quad (18)$$

Combining equation (17) and (18) now we have 5 polynomial constraint equations of degree 2 in the variables $\beta_1, \ldots, \beta_k$. Note that this approach applies even when the combined system is under constraint for example when the number of constraint equations are 8 or 10 from a limited number of points and/or line correspondences.

Case $N = 1$: The eigenvector corresponding to the lowest eigenvector is considered as the solution. The rotation matrix is constructed from the first nine element of the eigenvector and the scale is fixed by setting the determinant as 1.

Case $N = 2$: Two singular vectors corresponding to the lowest two eigenvalues are considered and the solution can be written as $\mathbf{x} = \beta_1 \mathbf{v_1} + \beta_2 \mathbf{v_2}$ where we need to solve for coefficients $\beta_1$ and $\beta_2$. Each of the five constraint equations consists of three monomials $[\beta_1^2, \beta_2^2, \beta_1\beta_2]$ and we solve for them using a linearization technique that was developed in cryptography and used in [21, 22]. The five constraint equations are arranged in a linear system of five equations of the form

$$\mathbf{A}\boldsymbol{\beta} = 0$$

(19)

Where $\mathbf{A}$ is the matrix constructed from the constraint equations and $\boldsymbol{\beta} = [\beta_1^2, \beta_2^2, \beta_1\beta_2]$. We solve this system using Singular Value Decomposition (SVD) and chose the signs of $\beta_i$ so that the points have positive z coordinates.

Case $N = 3$: We consider the three singular vectors corresponding to 3 smallest eigenvalues and the solution is written as

$$\mathbf{x} = \beta_1 \mathbf{v_1} + \beta_2 \mathbf{v_2} + \beta_3 \mathbf{v_3} \quad (20)$$

For ease of computation we use the 6 constraint equations in (16) and (17) and construct a linear system of 6 equations of the form

$$\mathbf{A}\boldsymbol{\beta} = \mathbf{b} \quad (21)$$

Now each of the constraint equations consists of 6 monomials $\boldsymbol{\beta} = [\beta_1^2, \beta_2^2, \beta_3^2, \beta_1\beta_2, \beta_1\beta_3, \beta_2\beta_3]$, $\mathbf{A}$ is a 6x6 square matrix constructed from the constraint equations and $\mathbf{b} = [0,0,0,1,1,1]^T$. We follow the same procedure as before to solve for $\beta_i$.

Case $N = 4$: We now have four unknown coefficients $\beta_i, i = 1, \ldots 4$ and the solution is written as

$$\mathbf{x} = \beta_1 \mathbf{v_1} + \beta_2 \mathbf{v_2} + \beta_3 \mathbf{v_3} + \beta_4 \mathbf{v_4} \quad (22)$$

In theory, the 5 constraint equations should still suffice. However, the linearization procedure treats all monomials in the constraint equations as unknowns and there are not enough constraints any more. We solve this problem by using a relinearization method used in [4] to solve for 4 unknown coefficients. Initially the degree of the system is reduced by dropping the scale. This is done by dividing each equations in (22) by $\beta_4$ to have a system of the form

$$\mathbf{x} = \beta_1' \mathbf{v_1} + \beta_2' \mathbf{v_2} + \beta_3' \mathbf{v_3} + \mathbf{v_4} \quad (23)$$

Now each of the 5 constraint equations consists of 10 monomials. Discarding the superscripts from the coefficients $\beta_i$, the monomials are $[\beta_1^2, \beta_2^2, \beta_1\beta_2, \beta_1, \beta_2, \beta_1\beta_3, \beta_2\beta_3, \beta_3, 1]$. In order to solve for a single variable, the constraint equations are rearranged to get a matrix system of the form

$$\begin{bmatrix} . & . & . & [1] & [1] & [2] \\ . & . & . & [1] & [1] & [2] \\ . & . & . & [1] & [1] & [2] \\ . & . & . & [1] & [1] & [2] \\ . & . & . & [1] & [1] & [2] \end{bmatrix} \begin{bmatrix} \beta_1^2 \\ \beta_2^2 \\ \beta_1\beta_2 \\ \beta_1 \\ \beta_2 \\ 1 \end{bmatrix} = \begin{bmatrix} 0 \\ 0 \\ 0 \\ 0 \\ 0 \end{bmatrix} \quad (24)$$

Where . represents some scalar value and $[N]$ denotes a polynomial of degree $N$ in the variable $\beta_3$. Once the matrix is constructed the next step is to solve for the null vector. The components of the null vector is obtained (up to a scale) by omitting the corresponding column of the 5x6 coefficient matrix, then taking the determinant of the resulting matrix

and multiply by ± as appropriate. For example, the $i^{th}$ component of the null vector can be computed as

$$u_i = (-1)^{i-1} \det(\hat{F}) \quad (25)$$

Where $\hat{F}$ is a 5x5 matrix constructed by deleting the $i^{th}$ column of the above 5x6 coefficient matrix. As a result, the components of the null vector $u = [\beta_1^2, \beta_2^2, \beta_1\beta_2, \beta_1, \beta_2, 1]$ introduces a new set of polynomial equations of degree 4 ($\beta_1^2, \beta_2^2, \beta_1\beta_2$), degree 3($\beta_1, \beta_2$) and degree 2(1) in the single variable $\beta_3$. The components of the null vector are not independent and they satisfy three constraint equations

$$\begin{aligned}(\beta_1) \times (\beta_2) &= (\beta_1\beta_2) \times (1) \\ (\beta_1) \times (\beta_1) &= (\beta_1^2) \times (1) \\ (\beta_2) \times (\beta_2) &= (\beta_2^2) \times (1)\end{aligned} \quad (26)$$

A set of three polynomial equations of degree 6 in the single variable $\beta_3$ are constructed by substituting the expressions for the components of the null vector from equation (19) in the three constraint equations in (20). Solving the nonlinear equation system (20) by linearization technique might lead to an inconsistent result from redundant equations and we approach it as an unconstrained optimization problem. Denoting the three univariate polynomials in equation (20) as $f_i(\beta_3); i = 1,2,3$ a cost function $F$ can be defined as a square sum of the three polynomials as $F = \sum_{i=1}^{3} f_i^2(\beta_3)$. The minima of $F$ can be identified by finding the roots of its 1st derivative $F' = \sum_{i=1}^{3} f_i(\beta_3) f_i'(\beta_3) = 0$. $F'$ is a 11th order polynomial which can be easily solved by the eigenvalue methods.

There can be up to 11 real roots of $F'(\beta_3)$, all of which need to be tested as a candidate solution. Once the candidate for $\beta_3$ are obtained, the corresponding candidates for $\beta_1, \beta_2$ are obtained by substituting values of $\beta_3$ in the component expressions of **u**.

Once the coefficients $\beta_i$ are computed with the method described above, the rotation can now be obtained by substituting $\beta_i; i = 1....k$ in equation (14). Then the scale is fixed by setting the determinant of the rotation matrix to 1.

Theoretically, the translation can also be found from equation (14). However, experiments with real data generated large translation error. Hence the translation is estimated separately in the following method once the rotation is estimated.

*E. Translation Computation*

In order to compute translation, we express the translation vector in terms of the rotation parameters in a least-square sense. We re-organize linear system of (23) in the form

$$\begin{bmatrix} \mathbf{A} & \mathbf{B} \end{bmatrix} \begin{bmatrix} \mathbf{r} \\ \mathbf{t} \end{bmatrix} = 0 \quad (27)$$

where **r** is the vector constructed from the elements of the rotation matrix and **t** is the translation vector

We rearrange equation (24) in the form $\mathbf{Ar} = -\mathbf{Bt}$ and then the least-square solution to the translation vector **t** can be expressed as

$$\mathbf{t} = -(\mathbf{B}^T\mathbf{B})^{-1}\mathbf{B}^T\mathbf{Ar} \quad (28)$$

*F. Efficient Computation*

In order to improve efficiency as well as to ensure numerical stability, computations of critical components of the algorithm are carefully designed. This way we ensure that the number of arithmetic operations does not become too large that might adversely affect the numerical behavior. We pre-compute the symbolic interaction between coefficients in the equation and then evaluate for values of the variables.

As an example, the coefficients of the matrix **A** in equation (14) are symbolically pre-computed and evaluated with the set of points and lines correspondences. Similar steps are taken for constructing systems of polynomials and solve them to estimate rotation and translation. The symbolic calculations are done using Matlab symbolic toolbox and the scripts are available at request. For brevity, details of the equations are skipped.

*G. Correct Pose Selection*

In order to find the correct camera, all the possible camera pose candidates obtained by the method in previous section need to be evaluated. We use a simple method to do this. For the corresponding points in first and second view, the 3D point is determined by back projecting the rays in first and second view. The 3D point is then projected onto the third view. A prescreening is performed by discarding the camera poses where majority of the 3D points have negative z coordinate in the 3rd camera before computing re-projection error. Similarly, lines back projected from the first and second image intersects at a 3D line. The 3D line is then projected onto the third camera and the line re-projection error is computed. The pose with the lowest re-projection error is selected as the final estimated pose.

IV. EXPERIMENTAL EVALUATION

The proposed algorithm is evaluated for accuracy, robustness and efficiency with both synthetic and real data. We demonstrate the effectiveness of the proposed algorithm by comparing results with state of the art line based pose estimation algorithms.

*A. Synthetic Experiments*

Monte Carlo simulation with synthetic lines and points were performed to quantify the performance of the proposed algorithm. We evaluate the algorithm's performance under various noise level, different lines and points' combination as well as large and small camera motion. At each trial, a predefined number of 3D line segments and 3D points are generated by randomly placing segment end points as well as the 3D points inside a cube (10 m length) that is centered at the origin of the world coordinate system $\{W\}$. We consider a virtual calibrated camera with image size 640x480 and focal length 800 pixels. The first view is generated by placing the camera randomly at a distance of 20 meters from the origin. For large motion, the second and third camera views are also generated randomly similar to the first view. For small motion, the second and third camera are generated by randomly selected camera positions that are within 5 meters of the first camera position. The cameras were then oriented so that they looked at the origin having all the 3D points and the line segments in the field of views.

The 3D points and line segments are projected on each view to generate the 2d points and line segments. Zero mean, Gaussian noise of varying standard deviations is added to the coordinates of the 2D points and line segments in each of the three views. All the experiments are based on 500 trials.

In the first experiment, we evaluate the effectiveness of the algorithm when used in a hypothesize-and-test framework such as RANSAC. It is important to select a small set of features for hypothesis generation in a RANSAC scheme, we used a combination of 3 or 4 lines and 3 or 4 points in this experiment. Similar to [1, 4], the lower quartile of the error distributions are used since finding a fraction of good hypothesis is more important than to get consistent results. Performance of the algorithm is shown in Fig. 3. The left image shows the camera pose as well as the lines and points. The relative distance between cameras could be as much as 40 meters. It can be seen that rotational and translational error increases with noise level. However with a moderate noise level, all combination of lines and points generate reasonable pose estimation. It can be observed that the pose estimation is more accurate when the number of points used is 4. This is explained by the fact that the number of constraint equations is larger in case of points than lines as well as how noise is added to the coordinates of the points and lines. For each line, random noise is added to both end points of the 2D lines resulting in a larger noise effect compared to 2D points. However, in moderate noise, the estimated camera poses are sufficiently accurate for further optimization.

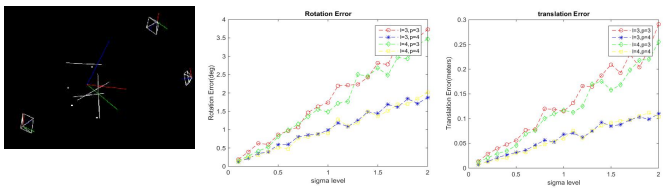

a. Experiment setup   b. Rotation Error   c. Translation Error

Fig. 3. Rotation and translation error with large camera motion as a function of noise level. Experiments are conducted with lines $n_L$ =3 or 4 and points $n_p$ =3 or 4.

In the second experiment, the robustness of the algorithm is evaluated by computing mean and median rotation and translation error as a function of noise level. The difference between mean and median error mainly comes from the pose estimation with large errors. The output of the algorithm is shown in fig. 4. The top row shows the mean rotation and translation error whereas the bottom rows shows the median errors. The experiments are done with 50% points and 50% lines where the numbers are varied from 4 to 8. It can be seen that the algorithm's performance is very stable when number of points and lines used are 5 or above. Again we consider large camera motion similar to experiment 1.

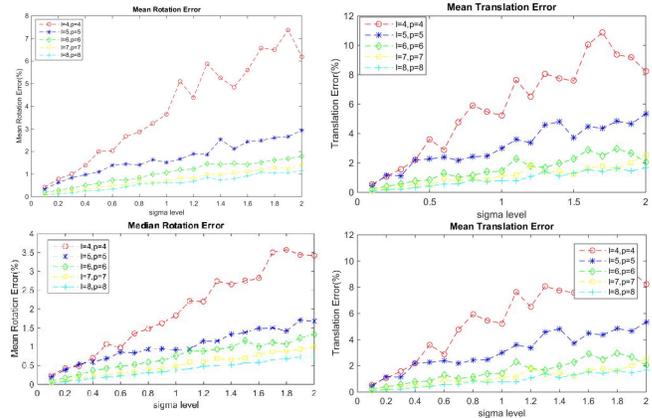

Fig 4. Mean and Median rotation and translation error as a function of noise level. Number of lines and points used ranges between 4 and 8. In each experiment, 50% lines and 50% points are used.

Efficiency of the method was evaluated on a laptop computer with an Intel Core i5-4340M 2.9 GHz CPU. Figure 5 shows the computational time requirement in millisecond vs the number of features. Number of lines and points are varied between 4 and 45. For 4 lines and 4 points the average runtime is 37 milliseconds and for 5 lines and 5 points, average runtime is 41 milliseconds. Results show that a combination of 4 points and 4 lines or 5points and 5 lines are suitable for use in RANSAC framework while being very stable.

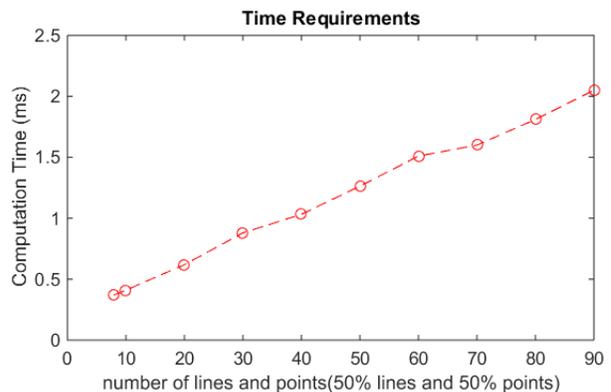

Fig. 5: Computational time requirement vs number of lines and points. All experiments are conducted with 50% lines and 50% points, averaged over 500 runs. The computational cost grows almost linearly with number of features used.

### B. Real Images

In order to evaluate the algorithm's performance in real world situations with various magnitude and direction of motions, we apply the algorithm on the VGG Multiview dataset.[22]. The dataset contains indoor and outdoor image sequences with extracted 2D lines and points, their reconstructed 3D positions and camera projection matrices.

The Corridor sequence contains indoor images with small forward camera motion whereas the other image sequences are outdoor images of buildings with various camera motions. In order to evaluate the accuracy and robustness of our algorithm we consider three images from each sequence and consider known camera poses for the first two views. Then we apply the algorithm to estimate the camera pose of the third view relative to the first view. The error is represented as the ratio of the estimated relative pose and ground truth relative pose between first and third view.

Fig. 6 shows the performance of the algorithm in boxplot representation, where each column depicts the distribution of errors for 100 runs of the algorithm for a number of points and lines. From left to right, the number of 2D lines and points combination considered are 2 lines + 2 points, 2 lines+ 3 points, 2 lines+4 points, 3 lines+3 points, 4 lines + 2 points, 4 lines + 4 points, 6 lines + 6 points, and 10 lines + 10 points. At each run, a predefined set of corresponding 2D lines and points in three views are randomly selected from the dataset. From the VGG Multiview dataset, the image sequences considered are UL= University Library, MC-I = Merton-College-I, MC-II=Merton-College-II, MC-III=Merton-College-III, WC=Wadham-College, C=Corridor.

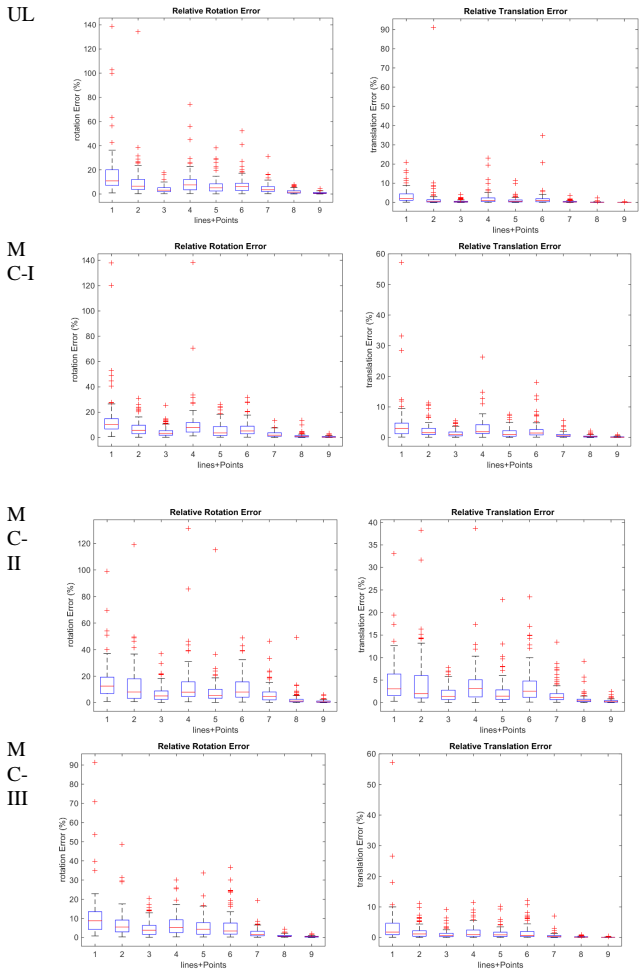
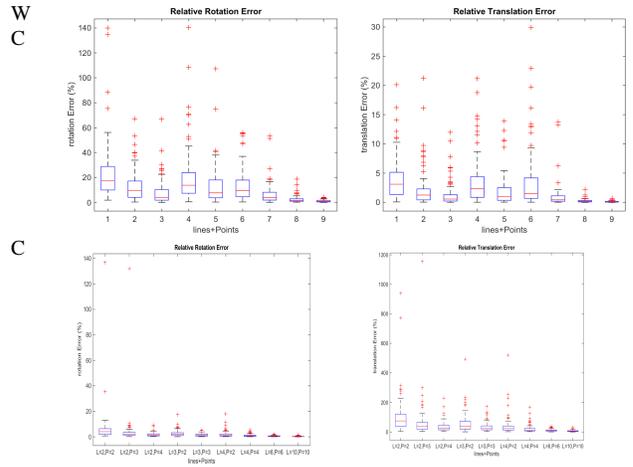

Fig. 6. Rotation and translation error as a function of number of lines and points. The boxplot result is generated by running the algorithm 100 times with randomly selecting $n$ lines and points. From left to right, the combinations of lines and points used are 2 lines + 2 points, 2 lines+ 3 points, 2 lines+4 points, 3 lines+3 points, 4 lines + 2 points, 4 lines + 4 points, 6 lines + 6 points, and 10 lines + 10 points.

It can be seen that the algorithm is very robust when using a combination of 3 lines + 3 points or above for both small and large camera motions. Figure 7 shows example results from the VGG dataset. In each row, the left and middle images are overlaid with the corresponding points and lines used for pose estimation of the third view relative to the first view. A set of 3D lines are then projected onto the third view using the estimated camera pose.

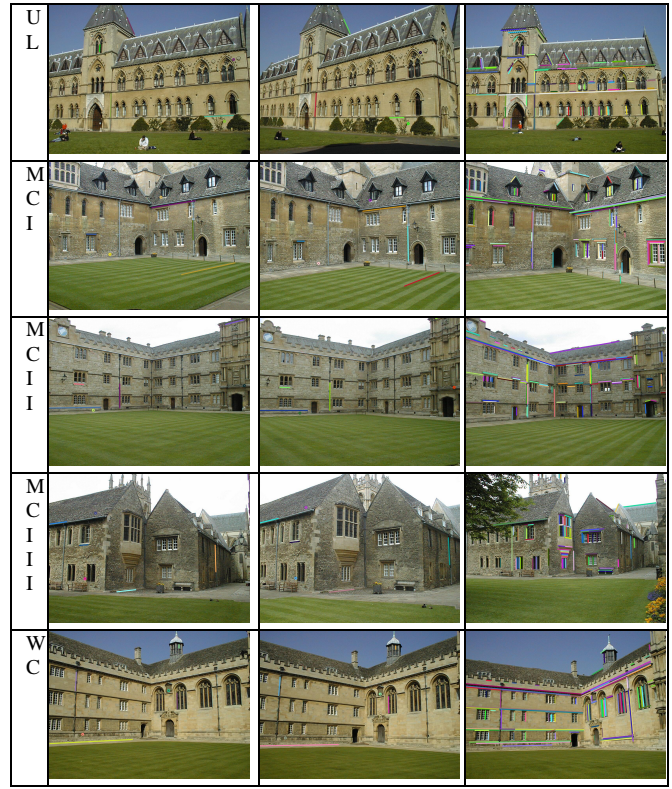

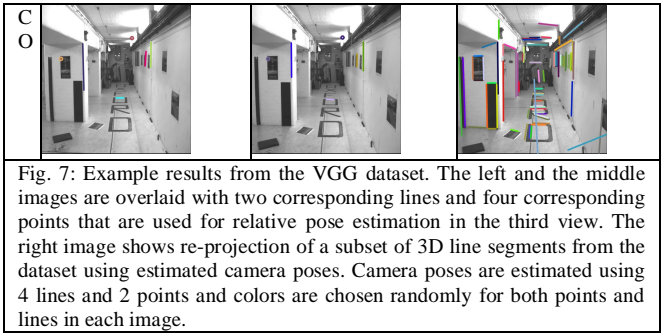

Fig. 7: Example results from the VGG dataset. The left and the middle images are overlaid with two corresponding lines and four corresponding points that are used for relative pose estimation in the third view. The right image shows re-projection of a subset of 3D line segments from the dataset using estimated camera poses. Camera poses are estimated using 4 lines and 2 points and colors are chosen randomly for both points and lines in each image.

## V. CONCLUSION AND DISCUSSION

In this work, we have presented the development of a unified relative camera pose estimation method that combines points and lines. Given a set of point and line correspondences in three views, the third camera view is estimated relative to the first view when the first two views are known. One of our primary contribution is a robust and efficient algorithm that is suitable for both small and large camera motions. We have demonstrated that the system, though not fully optimized, perform robustly and suitable for use in a RANSAC framework in both indoor and outdoor environment.

While no nonlinear optimization was performed in any of the experiments, it will conceivably improve the pose estimation. Future work include applying the algorithm in real time SFM pipeline and simultaneous localization and mapping frameworks.